\documentclass[10pt, a4paper]{article}

\usepackage{lrec-coling2024} 
\usepackage{times}
\usepackage{latexsym}
\usepackage{hyperref}
\usepackage{collcell}
\usepackage{booktabs}
\usepackage{tabularray}
\usepackage{tabularx}
\usepackage{multirow}
\usepackage{makecell}
\usepackage{multicol}
\usepackage{graphicx}
\usepackage{subcaption}
\usepackage{csquotes}
\usepackage{xcolor}
\usepackage{listings}

\title{When Simple Becomes Challenging: Adversarial Attacks based on Text Simplification Corpora}
\title{Simpler becomes Harder:\\ Is LLMs Behavior Consistent on Normal and Simplified Corpora?}
\title{Simpler becomes Harder:\\ Do LLMs Maintain a Coherent Behavior on Simplified Corpora?}
\title{Do LLMs Exhibit a Coherent Behavior on Simplified Corpora?}
\title{Simpler becomes Harder: Do LLMs Exhibit a Coherent Behavior on Simplified Corpora?}

\name{Miriam Anschütz, Edoardo Mosca, Georg Groh} 

\address{Technical University of Munich \\
         School of Computation, Information and Technology \\
         \{\href{mailto:miriam.anschuetz@tum.de}{miriam.anschuetz}, edoardo.mosca\}@tum.de, grohg@cit.tum.de\\}

\abstract{
Text simplification seeks to improve readability while retaining the original content and meaning. Our study investigates whether pre-trained classifiers also maintain such coherence by comparing their predictions on both original and simplified inputs. We conduct experiments using 11 pre-trained models, including BERT and OpenAI's GPT 3.5, across six datasets spanning three languages. Additionally, we conduct a detailed analysis of the correlation between prediction change rates and simplification types/strengths. Our findings reveal alarming inconsistencies across all languages and models. If not promptly addressed, simplified inputs can be easily exploited to craft zero-iteration model-agnostic adversarial attacks with success rates of up to 50\%.
 \\ \newline  }

\begin{document}

\maketitleabstract

\section{Introduction}
Automatic text simplification (ATS) is a popular natural language processing task that creates texts in plain language, preserving the original message of the source text. Plain or simplified language is a version of the English language with reduced text complexity and uses only well-known vocabulary. This simplified version aims to increase accessibility and, thus, gives people with learning impairments or reading difficulties access to information on the internet \citep{martin-muss}. 
While simplifications must alter some text features to reduce its overall complexity, they should still preserve the original source's content. Indeed, content coherency between the original source and simplified output is a core element of text simplification, spanning across various aspects (sentiment, emotion, topic, etc.) \citep{saggion-simplification-meaning}. For instance, if a strong sentiment or emotion, e.g., anger about something, is conveyed in the original text, this emotion should also be perceivable in the simplified version as well.

In line with this thought, this paper investigates whether models also exhibit this coherent behavior and assesses pre-trained classifiers and recent large language models (LLM) like GPT3.5 on original and simplified texts. For this, we exploit text simplification corpora across different languages, let the models classify content-related features such as the addressed topic, emotion, or sentiment, and analyze potential variations in these labels. 
\begin{figure}[ht]
    \centering
    \includegraphics[width=\linewidth]{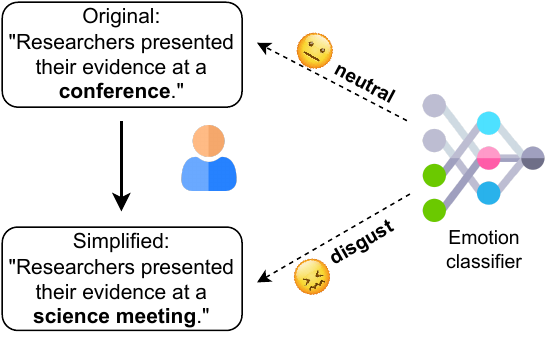}
    \caption{Manually created sentence pair. The simplified version simplifies the word \enquote{conference} but preserves the meaning and neutral sentiment of the original sentence. However, a pre-trained emotion classifier behaves incoherent and predicts a different label for the simplified sentence.}
    \label{fig:vis_abstract}
\end{figure}

Our results show that models change their predictions for up to 50\% of the samples, depending on the language and task used. \autoref{fig:vis_abstract} shows an example of an incoherent model behavior on a manually created sample. The simplified version replaces \enquote{conference} with \enquote{science meeting}. Apart from that, both versions convey the same message and sentiment. Nevertheless, a pre-trained emotion classifier assigns different labels to the two samples. By using human-created or human-aligned simplification corpora, we can ensure that our benchmark samples are natural and that humans consider them meaning-preserving and valid alterations. 

Therefore, our contribution can be summarized as follows:
\begin{itemize}
    \item We compile a strict selection of human-created or human-aligned simplification datasets as model consistency benchmarks. This ensures the naturalness and correctness of our benchmark samples.
    \item We test pre-trained classifiers for their prediction consistency on normal and simplified language, covering multiple tasks across different languages to give extensive insight into the models' shortcomings.
    \item Results show concerning discrepancies between the model behavior on normal and simplified inputs. If not addressed, these discrepancies can be easily exploited to produce zero-iteration model-agnostic adversarial attacks with a success rate of up to 50\%.
\end{itemize}

\section{Background and related work}
Previous work by \citet{elazar-consistency} defined model consistency as follows: Given two equivalent paraphrases, a consistent model creates non-contradictory predictions for both of them. In their study, they probed language models with cloze-phrases and evaluated whether the predictions for the original and a paraphrase prompt were similar. The models produced contradicting predictions for 39\%-51\% of the samples, depending on the model.
We extend the consistency definition by \citet{elazar-consistency} to classification tasks and deem a model consistent if it assigns the same label to both versions.

Our work is inspired by model robustness checks with adversarial attacks. To create an adversarial attack, a pre-trained classification model $f$, text samples $x_i$, and model predictions $f(x_i)=y_i$ are given. Then, the attacker tries to find adversarial samples $x'$ that fool the model to change its prediction compared to the original sample, hence $f(x_i')=y_i' \neq y_i$. The changes to the original sample $x$ should be minimal and not change its ground truth. In addition, humans should perceive the alteration as valid and natural \citep{qi-style-attack}. If a model changes its prediction for the sample $x'$, it is considered sensitive to the specific adversarial attack.

Approaches to adversarial attacks can be classified based on their knowledge of the target model. In a \textit{white-box} scenario, the attacker has full access to the model and can optimize perturbations using its output and gradients to find adversarial samples. \textit{Black-box} attacks, on the other hand, only have access to predictions without further model information \citep{yoo-searching}. \textit{Blind or zero-iteration} attacks lack any model feedback and can only apply one perturbation step to deceive the model. This study adopts a zero-iteration setting to assess model sensitivity to text simplification perturbations.

Similar to our objective, \citet{van-etal-simplification-preprocessing} examined how NLI models change their predictions when the samples are pre-processed by an automatic simplification model and observed a performance drop of up to 50\%. We extend this research to further tasks and languages. Another study investigated whether models are robust to text-style transfer attacks. 
\citet{qi-style-attack} utilized a pre-trained style transfer model to transfer common datasets for sentiment or topic prediction, e.g., into Twitter, bible, or poetry language style. In most cases, at least one style adaption yielded the model to alter its prediction. However, a human survey found that many transfers altered the ground truth of samples, indicating that changing predictions was appropriate behavior.
To avoid unnatural paraphrases that change the samples' ground truth, we rely on human-supervised datasets for model coherence checks instead of generating adversarial samples with pre-trained simplification models.


\section{Methodology}
This paper examines whether models maintain a consistent behavior when dealing with normal and simplified texts. For this, we investigate whether and to what extent human-created simplification datasets can lead to output changes in pre-trained models published to the Hugging Face model hub \citep{Wolf-Huggingface}. We assume that although the altered text style is perceptible to humans, it is still a natural and meaning-preserving paraphrase of the original. 
Simplifications are targeted toward people with lower reading understanding capabilities. As such, the simplifications should reduce the complexity of the text but still preserve the topic or sentiment conveyed in the original texts. We expect simplification corpora, used to train ATS systems, to reflect this aspect.
 
In our experiments, we test various classification tasks covering topic, emotion, fake news/toxicity, and sentiment prediction. For all of them, we expect the original and simplified text to have the same content-related features and, thus, get the same labels from the classifiers. The specific tasks vary across languages, depending on the availability of pre-trained models for the respective language. In the following, we introduce the models under test, our selected datasets, and their pre-processing.

\subsection{Models}
\begin{table*}[ht]
    \centering
    \begin{tabularx}{\textwidth}{p{2.8cm}Xccc}\toprule
        \textbf{Prediction task} & \textbf{Model} & \textbf{\#Classes} & \textbf{Domain} & \textbf{Benchmark} \\ \midrule
        \multicolumn{4}{c}{\textbf{English}}\\ 
        Topic &bert-agnews \citeplanguageresource{lee-ag-news-model}& 4 & news &94\%*\\ 
        Sentiment & bert-base-multilingual-uncased-sentiment \citeplanguageresource{su-sentiment-model} & 5 & online & 67\%*\\ 
        Emotion & emotion-english-distilroberta-base \citeplanguageresource{hartmann-emotion-model} & 7 & diverse & 62\%*\\ 
        Fake news & roberta-fake-news-classification \citeplanguageresource{benyamina-fake-news-model}& 2 & news & 100\%*\\ 
        \midrule
        \multicolumn{4}{c}{\textbf{German}}\\
        Topic & bert-base-german-cased-gnad10 \citeplanguageresource{laiking-topic-gnad10} & 9 & news & 68\%*\\
        Sentiment & german-news-sentiment-bert \citeplanguageresource{lüdke-sentiment-news} & 3 & news & 96\%*\\
        Toxicity & distilbert-base-german-cased-toxic-comments \citeplanguageresource{Buschmeier-toxicity} & 2 & social media& 78\%*\\
        \midrule 
        \multicolumn{4}{c}{\textbf{Italian}}\\
        Topic & it5-topic-classification-tag-it \citeplanguageresource{Papucci-topic} & 10 & online& 57\%\\
        Sentiment & feel-it-italian-sentiment \citeplanguageresource{bianchi-feel-IT} & 2 & twitter&84\%*\\
        Emotion & feel-it-italian-emotion \citeplanguageresource{bianchi-feel-IT} & 4 & Twitter& 73\%*\\ 
        \bottomrule
    \end{tabularx}
    \caption{Pre-trained classifiers used to perform different classification tasks. The classifiers vary in their number of classes and training data domain.\newline * performance copied from the Hugging Face model page}
    \label{tab:models}
\end{table*}
\autoref{tab:models} shows the pre-trained classification models we selected from the Hugging Face model hub \citep{Wolf-Huggingface}. Not all classification tasks are suited for our experiments as some text features in the simplified versions, for example, the complexity, are altered on purpose. Therefore, we selected only content-related prediction tasks like topic or sentiment classification. 
Furthermore, we picked models with varying numbers of classes to experiment with different task difficulties and tried to reflect the datasets' domains to avoid misclassifications due to domain mismatch. However, we could not always find models with matching domains and, thus, preferred a mismatching domain over skipping the task in the respective language.

As expected, the English language has the highest availability of pre-trained models, and we found models with matching domains for all four tasks. For German, we found in-domain classifiers to predict the sentiment and topic but not for fake news or emotion prediction. To counteract the low availability of suited models, we included a toxicity classifier that detects toxic comments on social media data. Finally, for Italian, we could only find an in-domain topic classifier and had to include an out-of-domain emotion and sentiment classifier.

As shown in \autoref{tab:models}, the models show different performances on relevant benchmark datasets. For most of the model evaluations, we used the test set accuracies reported in the respective model pages on the hub. Only for the Italian topic classifier, no score was reported. Therefore, we evaluated the model on the TAG-it test set \citep{cimino-TAG-it} ourselves.
\subsection{Datasets}
A simplification corpus contains texts in standard language aligned with their simplified version. These texts can be sentences or paragraphs, and the simplifications span from replacing single words to completely rewriting the text.

While there exists a collection of simplification corpora across many languages \citep{ryan-MultiSim}, we had specific requirements for our benchmark datasets. First, we selected only datasets where humans created the texts or alignments to avoid label changes due to misalignments. Second, a paragraph can address different topics and have multiple sentiments, resulting in ambiguous classifications. Hence, we only selected sentence-level datasets where the simplifications had a large enough overlap with the original version. In addition, we restricted ourselves to corpora with multiple levels of simplifications to examine whether the strength of the simplification had an impact on the prediction change rates. Moreover, the availability of pre-trained classification models limited the language diversity. Finally, we compiled a benchmark dataset for investigations in English, German, and Italian. 
\begin{table*}[ht]
    \centering
    \begin{tabularx}{\linewidth}{p{6.5cm}Xcr}\toprule
        \textbf{Dataset}& \textbf{Domain} & \makecell[c]{\textbf{\#Simplification}\\\textbf{levels}} & \textbf{\#Samples}  \\ \midrule
        \multicolumn{4}{c}{\textbf{English}}\\
        Newsela EN \citeplanguageresource{xu-newsela}& news& 4 & ~61k\\ 
        \midrule
        \multicolumn{4}{c}{\textbf{German}}\\
        TextComplexityDE \citeplanguageresource{naderi-textComplexityDE}& wikipedia & 2* & 249 \\ 
        DEplain \citeplanguageresource{stodden-deplain} & online & 2 & 1.846\\ 
        \midrule
        \multicolumn{4}{c}{\textbf{Italian}}\\
        Simpitiki \citeplanguageresource{tonelli-simpitiki}& wikipedia & 1 & 1.163 \\ 
        AdminIT \citeplanguageresource{miliani-admin-it}& wikipedia, government & 2* & 736 \\ 
        Terence/Teacher \citeplanguageresource{brunato-teacher-corpus}& online, literature & 2* & 1.146 \\ 
        \bottomrule
    \end{tabularx}
    \caption{Simplification datasets used to retrieve adversarial data with their number of simplification levels and covered domains. Datasets with the number of simplification levels marked with a $^*$ differ from their original version.}
    \label{tab:datasets}
\end{table*}

\autoref{tab:datasets} shows the datasets we selected for your study. For English, we used Newsela \citeplanguageresource{xu-newsela}, the gold standard for English simplification \citep{martin-muss}: This dataset was created by language experts and consists of sentences from news articles that were simplified into four different levels, where \textit{V1} is the mildest and \textit{V4} the strongest simplification.

For German, we selected multiple datasets. The TextComplexityDE dataset by \citetlanguageresource{naderi-textComplexityDE} consists of sentences from Wikipedia and their simplified versions. The simplifications were obtained from native speakers and are annotated by their simplification strength. Initially, the dataset has three simplification levels. However, we discarded the sample annotated with \enquote{could not be simplified}, resulting in the simplification strengths \textit{slightly simplified} and \textit{strongly simplified}.
The second corpus, DEplain \citeplanguageresource{stodden-deplain}, is compiled from online articles with document- and sentence-level alignments. We picked the test split of the sentence-level dataset. The samples in the dataset were aligned manually and were annotated by their CEFR language level \citep{council-cefr}. The original samples are at level B2 or C2 and simplified into A1 or A2. To match the simplification levels of the TextComplexityDE data, samples with an original language level of B2 are considered \textit{slight simplifications}, while samples with original level C2 are \textit{strong simplifications}. 

We investigate three different simplification corpora for Italian. Simpitiki \citeplanguageresource{tonelli-simpitiki} uses the edit history of Italian Wikipedia to select edits with the annotation \enquote{simplification}. The authors manually labeled the samples by their simplification operation and removed non-simplified versions. The second corpus, AdminIT \citeplanguageresource{miliani-admin-it}, contains a subset of the Simpitiki data and sentences from Italian municipality homepages. The samples are categorized into three simplification strategies: samples with the label \textit{OP} show only a single simplification operation like sentence split or lexical substitution, while other samples were either manually rewritten (label \textit{RS}) or manually aligned based on simplified documents (label \textit{RD}). We grouped the later categories together, yielding two levels of simplifications present in the corpus: single simplification operation and more complex rewritings.
The final dataset consists of two subcorpora, the Terence and the Teacher corpus \citeplanguageresource{brunato-teacher-corpus}. Terence is created from books for children that were simplified manually. In contrast, the Teacher corpus contains original/simplified texts from educational homepages for teachers. Both datasets were manually annotated by their simplification operations. We divided the total number of annotations per text by the number of sentences and grouped the samples into two simplification levels, one with one or fewer simplification operations per sentence and one with multiple.

\section{Evaluation}
\begin{figure*}
    \begin{subfigure}[b]{0.32\textwidth}
        \centering
        \includegraphics[width=\textwidth]{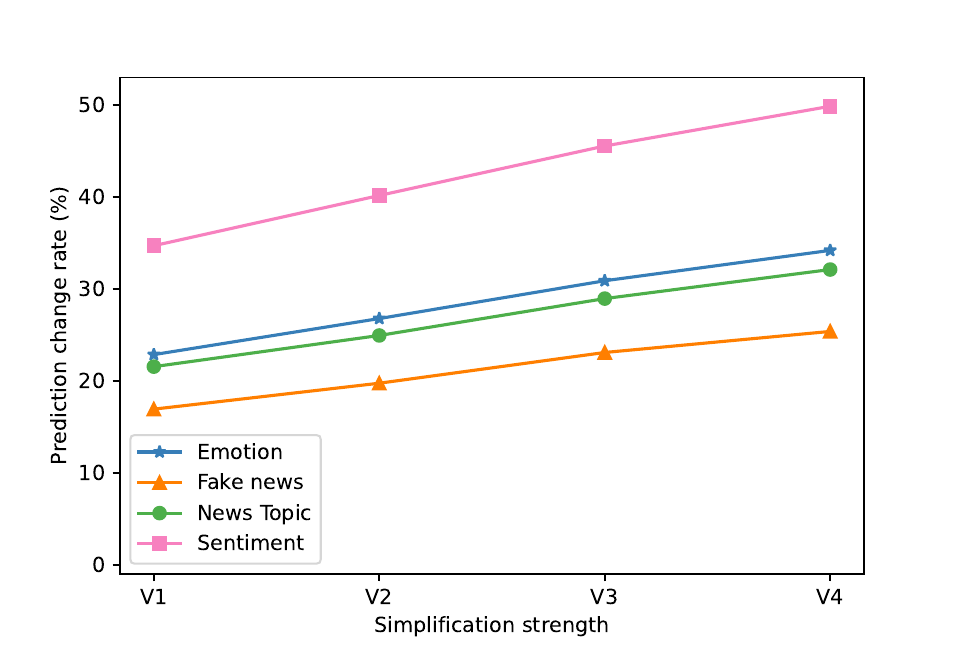}
        \caption{English}
    \end{subfigure}
    \hfill
    \begin{subfigure}[b]{0.32\textwidth}
        \centering
        \includegraphics[width=\textwidth]{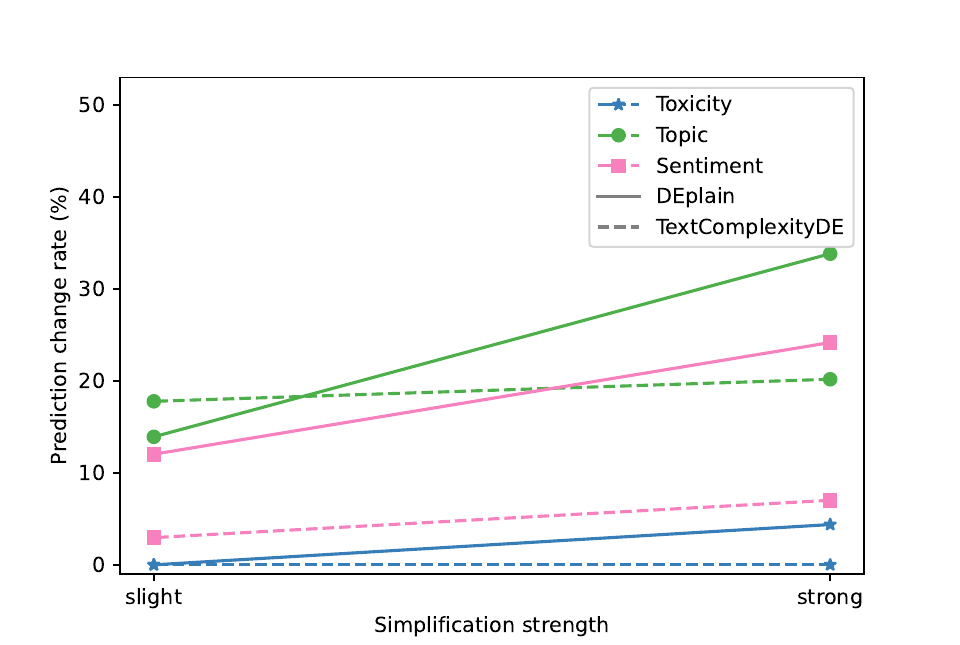}
        \caption{German}
    \end{subfigure}
    \hfill
    \begin{subfigure}[b]{0.32\textwidth}
        \centering
        \includegraphics[width=\textwidth]{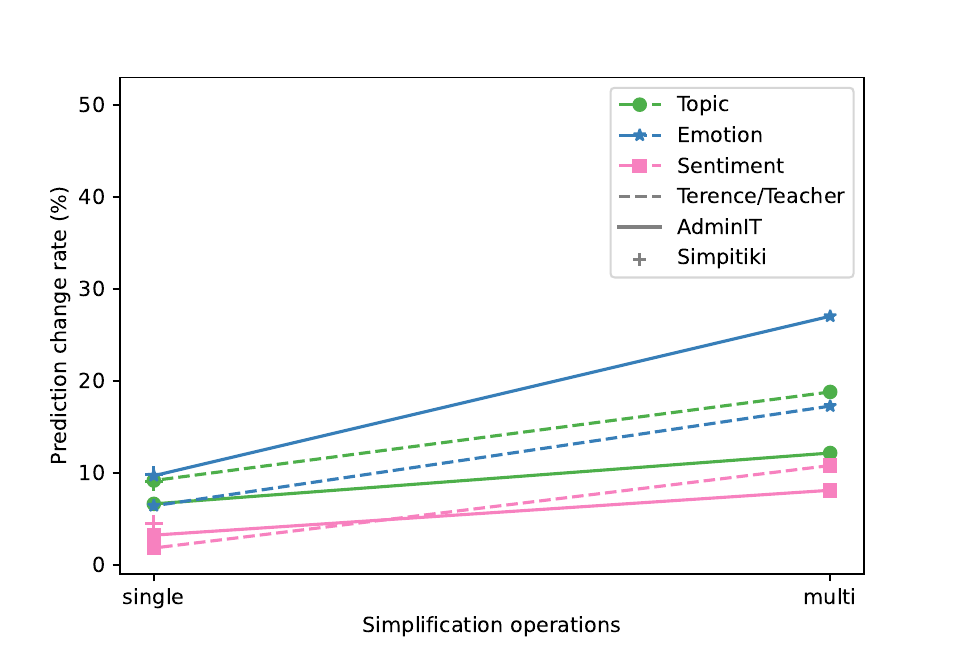}
        \caption{Italian}
    \end{subfigure}
    \caption{Prediction change rates across different languages and tasks, sorted by the simplification strength of the samples. All classifiers show more deviating predictions the stronger the simplification strength. Overall, the English models are least coherent.}
    \label{fig:asr_langugages}
\end{figure*}
For each sample in the text simplification corpora, we obtained each classifier's prediction for the original and the simplified text and compared their predictions. For our consistency analysis, we do not evaluate whether any of the classifiers predicts a wrong label. We expect the classifiers to label all samples the same, whether they are the original or simplified versions. If the predictions deviate, we 
consider the respective classifier inconsistent with these samples.
We then counted the number of samples with deviating predictions and compared the counts to the full dataset size to obtain the prediction change rate (PCR) for each classifier. \autoref{fig:asr_langugages} shows the PCRs for all classifiers across different languages. We observe rates around 20\% on average and up to 50\% for English Newsela. These change rates are high, especially considering that the simplified samples are created without any knowledge of the models and can thus be considered model-agnostic adversarial samples. They are used for all classifiers simultaneously. Other works that limited the number of model-specific changes to the sample achieved only prediction change rates below 10\% \citep[Fig. 2]{yoo-searching}.

A common trend across all languages is that the model is more likely to change its prediction the stronger the simplification is---i.e., the more simplification operations are performed. Comparing the topic classifiers among all languages (green curves in \autoref{fig:asr_langugages}), the English classifier has four classes with an accuracy of 94\% and is sensitive to simplification in more than 30\% of the time, while the Italian topic classifier has ten classes with an accuracy of only 57\% that only shows deviating predictions for 10\% of the samples. This indicates the number of labels and the classifier's performance seems to have little impact on the prediction change rate. Similarly, the Italian models have a strong domain mismatch (Wikipedia data vs. Twitter-trained models), while the English data and most of the models are from the news domain. Nevertheless, the English models are more easily affected by simplified inputs. Overall, the human-created or human-aligned samples in the different simplification corpora evoke an alarming amount of prediction changes.

In the following sections, we discuss further experiments to investigate factors influencing classifiers behavior (sections \ref{sec:edit_distance} - \ref{sec:masking_named_entities}) as well as examine whether s.o.t.a. LLMs like OpenAI's GPT models \citep{openai-gpt4} are also sensitive to simplifications (section \ref{sec:chatgpt}).
\subsection{Edit distances}\label{sec:edit_distance}
\begin{figure}[t]
    \centering
    \includegraphics[width=\linewidth]{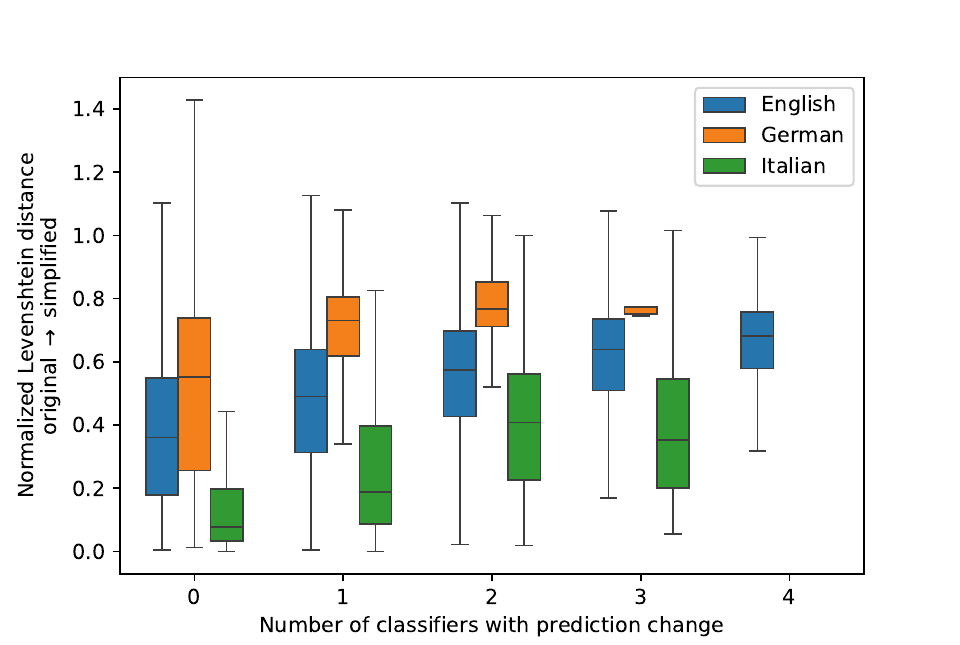}
    \caption{Number of classifiers with changing predictions per sample and their Levenshtein distances between the original and simplified sentences. The distances were normalized by the sample's lengths.}
    \label{fig:lev_distances}
\end{figure}

As stated before, the prediction change rates increase with a higher level of simplification. An obvious assumption would be that this is due to increasing differences between the original and simplified samples, especially since higher-level simplifications sometimes remove parts of the original information. To verify this assumption, we calculated the Levenshtein distances, normalized by the original sample's lengths, between the original and simplified versions using the Python Levenshtein\footnote{\url{https://rapidfuzz.github.io/Levenshtein/levenshtein.html\#distance}} package. In \autoref{fig:lev_distances}, these ratios are correlated with the number of classifiers that changed their prediction for the respective sample. Some samples have a normalized distance larger than $1$, e.g., when an explanation is added in the simplified version. Among all languages, the German samples have the highest ratios.
For all languages, the samples with no prediction change have the smallest normalized distances. However, the average distance for samples with two or more classifiers with prediction changes stays the same or even decreases. Overall, we observe a Spearman correlation of $0.34$ for English, $0.35$ for German, and $0.34$ for Italian between the normalized Levenshtein distance and the number of classifiers with changing predictions\footnote{p-values are $0.0$ (en), $8.5\mathrm{e}{-65}$ (de), and $1.4\mathrm{e}{-85}$ (it); calculated with SciPy: \url{https://docs.scipy.org/doc/scipy/reference/generated/scipy.stats.spearmanr.html}}.
Therefore, only a weak correlation exists between the normalized distance and coherency of the models. This indicates that the simplification operation and the choice of vocabulary to simplify the samples are more relevant to the classifiers than the pure number of edit operations or parts removed from the original sentence.

\subsection{Reducing task complexities}
\begin{figure}[ht]
    \centering
    \includegraphics[width=\linewidth]{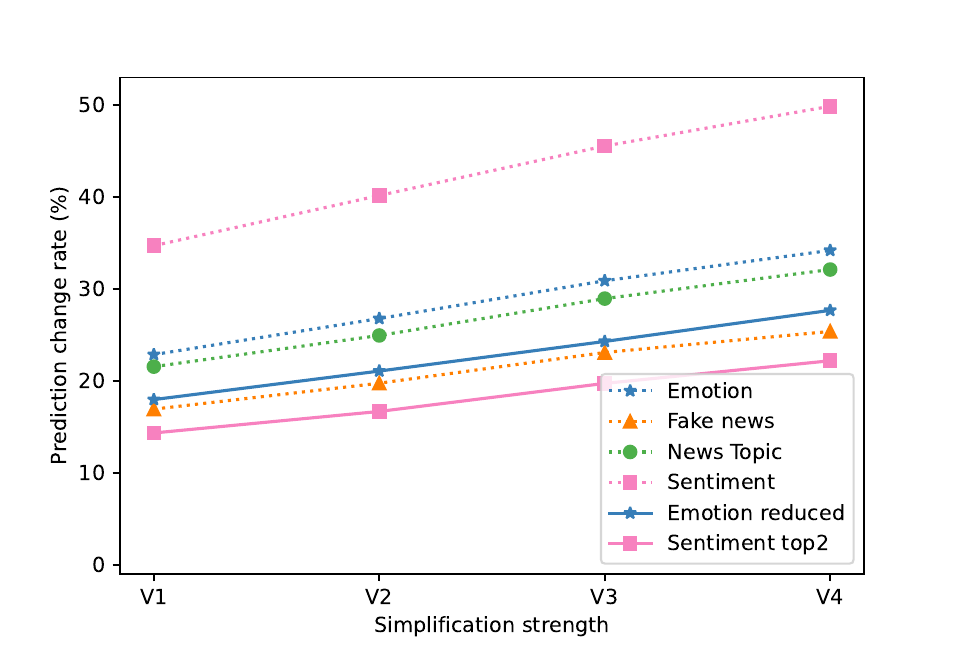}
    \caption{Predictions change rates for tasks with reduced number of classes. The reduced tasks have a better performance but are still susceptible to simplifications.}
    \label{fig:tasks_reduced}
\end{figure}
Fine-grained emotion and sentiment prediction tasks can be difficult and may induce some ambiguity for the models. The English emotion and sentiment classifiers have seven and five classes, respectively. To control for these potential ambiguities, we reduced the number of classes and, thus, the task complexities. For the emotion task, we summarized all negative emotions, like anger, disgust, fear, sadness, and surprise, into a negative class. The only positive class, joy, and the neutral class were kept, resulting in a three-class classification task with which we can detect emotion flips. For the sentiment task, we calculated the top-2 accuracy. That is, a prediction was only considered as a deviating prediction if the difference was at least two steps (e.g., strongly negative vs. negative is no prediction change, but strongly negative to neutral is). \autoref{fig:tasks_reduced} shows how the classifier predictions change for these reduced tasks compared to the original tasks. The prediction change rate drops by five percentage points for the emotion tasks and more than 20 percentage points for the sentiment task. The rate is still higher with increased simplification strength. We conclude that the difficulty and ambiguity of the classification task can influence the prediction change rate. Nevertheless, even with the reduced tasks, the classifiers still perform inconsistently with simplified versions. 

\subsection{Simplification operations}
\begin{figure}[ht]
    \centering
    \includegraphics[width=\linewidth]{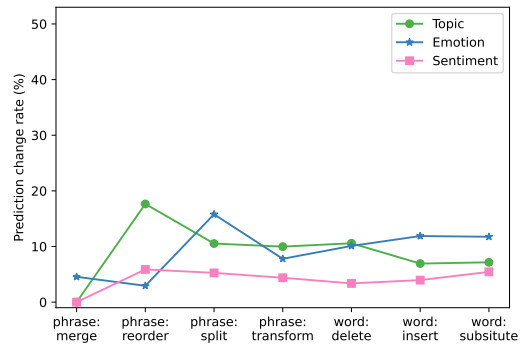}
    \caption{Prediction change rate of different simplification operations as annotated in the Italian Simpitiki corpus \protect\citeplanguageresource{tonelli-simpitiki}.}
    \label{fig:simp-ops}
\end{figure}

We further investigated which simplification operations especially tempt the classifiers to change their predictions. For this, we utilized the Italian Simplitiki corpus \citeplanguageresource{tonelli-simpitiki}. The samples in this corpus are annotated by the operation performed to obtain the simplified version. These operations can be on the word level, such as deletion or replacement of single words, or on the phrase level, for example, splitting a sentence into two or transforming the verbal voice. Samples with word-level changes are closer to their original than phrase-level ones. Therefore, we expected the word-level operations to result in lower change rates. However, \autoref{fig:simp-ops} shows that they are on par or lead to even more prediction changes than the phrase-level simplifications. Word substitution is a combination of word deletion and insertion and, as such, has the highest PCR of all word-level perturbations. As such, replacing a word with its synonym has been used as word-level adversarial attacks before \citep{chiang-word-synonym}.
On the phrase level, merging two sentences does not affect the sentiment and topic classifiers. Yet, the topic classifier is sensitive to the information order, exhibiting the highest PCR of almost 20

\subsection{Masking named entities} \label{sec:masking_named_entities}
Named entities (NE) can strongly impact the sentiment or topic of a phrase as, sometimes, they only occur in a particular context. In simplified language, NEs are sometimes generalized or removed. To test which influence named entities have on the models' prediction consistency, we compared the predictions when masking the named entities.

We utilized Spacy \citep{Honnibal-spacy} to detect named entities in our original and simplified phrases. We searched for tokens with a tag in this list: ['EVENT', 'GPE', 'LANGUAGE', 'LAW', 'LOC', 'NORP', 'ORG', 'PERSON', 'PRODUCT', 'WORK\_OF\_ART']. If such a token was found, we masked it by replacing it with the placeholder \enquote{NAME}. With this, the NE in the aligned pairs are the same and do not impact the classification outcome.
\begin{figure}[ht]
    \centering
    \includegraphics[width=\linewidth]{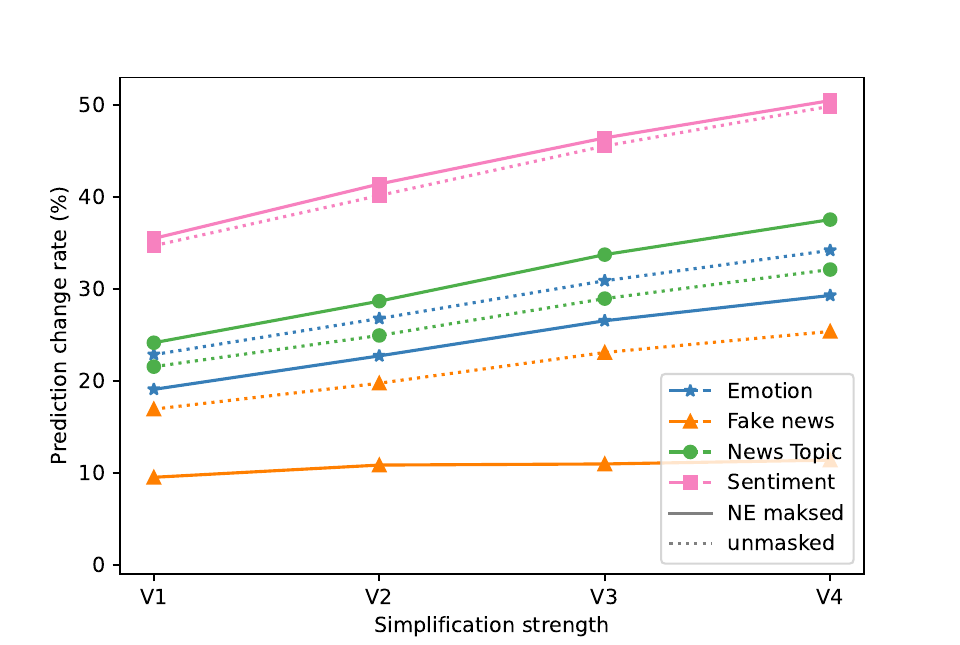}
    \caption{Comparison of prediction change rates of masked (solid) and unmasked (dashed) named entities on the English Newsela corpus. It depends on the task if NE masking increases or decreases the PCR.}
    \label{fig:ne-masked}
\end{figure}

In \autoref{fig:ne-masked}, we compare the performances of the classifiers on the named-entity-masked and unmasked data. For the fake news and emotion classification tasks, masking the named entities reduces the prediction change rates. In contrast, for the topic prediction task, the masking even increases the change rates, probably because the sentences become more ambiguous. Therefore, it depends on the classifier and its task whether the possibly changing NEs in the simplifications impact its consistency.

\subsection{ChatGPT} \label{sec:chatgpt}
OpenAI's GPT models \citep{openai-gpt4} are among the NLP models displaying the strongest capabilities in terms of generalization and performance. Hence, we also examined whether these models are sensitive to our simplifications. We used the OpenAI API to query the \texttt{gpt-3.5-turbo} model and predicted samples from the Newsela dataset in a one-shot manner. We prompted each sample individually to avoid biases due to previously seen non-simplified versions. We probed the same tasks with the same labels as the English models from the Hugging Face model hub described above. In addition, we asked the model to return a dictionary with all predictions simultaneously and set the temperature parameter to zero. The full prompt can be found in \autoref{fig:gpt_prompt}. Due to the long processing time of an API request and financial limitations, we restricted ourselves to English and only classified the first 1000 samples per simplification level.

\definecolor{darkgreen}{HTML}{2f9e4d}
\lstset{language=Python,basicstyle=\small\tt, breaklines=true,breakatwhitespace=true, columns=fullflexible, showstringspaces=false,
escapeinside={*@}{@*},
emph={"role":, "content":, "name"},emphstyle={\color{blue}},
keywordstyle={[3]\ttfamily\color{darkgreen}},stringstyle=\color{white!20!black},
}
\begin{figure}
    \begin{lstlisting}
{*@\color{darkgreen}"role"@*: "system", *@\color{darkgreen}"content"@*: "You are an assistant designed to label news texts. Users will paste in a string of text and you will respond with labels you've extracted from the text as a JSON object. The topic must be one of world, sports, buisiness or sci/tech. The sentiment is on a scale from 1 to 5 stars. Fake news can be true or false. Emotion can be one of anger, disgust, fear, joy, neutral, sadness or surprise"},
{*@\color{darkgreen}"role"@*: "system", *@\color{darkgreen}"name"@*: "example_user", *@\color{darkgreen}"content"@*: "Predict the sentence: Even a big first-day jump in shares of Google (GOOG) couldn't quiet debate over whether the Internet search engine's contentious auction was a hit or a flop."},
{*@\color{darkgreen}"role"@*: "system", *@\color{darkgreen}"name"@*: "example_assistant", *@\color{darkgreen}"content"@*: "{
        'topic': 'sci/tech', 
        'sentiment': '2 stars',
        'fake_news': False,
        'emotion': 'sadness'}"},
{*@\color{darkgreen}"role"@*:"user", *@\color{darkgreen}"content"@*: f"Predict the sentence: {sentence}"}
    \end{lstlisting}
    \caption{OpenAI system description for our one-shot classification approach.}
    \label{fig:gpt_prompt}
\end{figure}

\begin{figure}
    \centering
    \includegraphics[width=\linewidth]{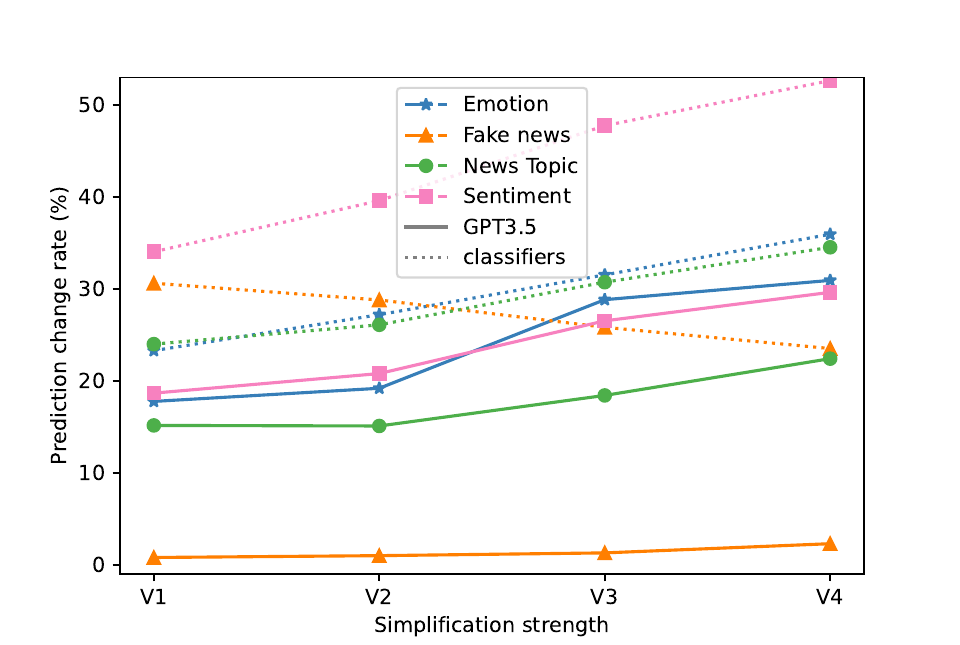}
    \caption{Comparison of Hugging Face classifiers with OpenAI's GPT 3.5. Especially for the emotion and the sentiment task, GPT is only slightly more robust than the smaller, task-specific models.}
    \label{fig:newsela-gpt}
\end{figure}

\autoref{fig:newsela-gpt} shows the prediction change rates on this subset. We compare different simplification levels and tasks for the ChatGPT model and our Hugging Face classifiers. The classification change rates for the fake news detection task decrease significantly compared to the model by \citetlanguageresource{benyamina-fake-news-model}. The PCRs for the other models decrease slightly, and the emotion classification is almost on par with the Hugging Face classifier in the strong simplifications. As we have seen before with the task-specific models, the changes increase with stronger simplification levels. Therefore, even ChatGPT is not robust to text simplification and makes incoherent predictions.

\section{Discussion}
This paper shows that even s.o.t.a models like GPT3.5 are sensitive towards text simplification as a special form of text style transfer and produce incoherent predictions for texts and their simplified versions.
We observe this behavior on human-generated or human-aligned text simplification datasets.
That means that our samples were validated by humans before and, thus, should be natural, grammatically correct, and meaning-preserving. However, previous work by \citet{devaraj-simplification-factuality} has shown that even human-curated simplification datasets can contain factual errors. While pre-processing and filtering those datasets seems promising for text simplification training \citep{ma-improving-factuality}, we worked with the original dataset versions. Therefore, it is possible that some of the labels change due to factual errors in the datasets. Unfortunately, such investigations only exist for English corpora. Assuming that the corpora in other languages are of higher factuality, this could explain why the English classifiers showed the highest prediction change rates even though English is usually the best-resourced language with top performance.

We tried to select well-documented classifiers with high accuracy rates. Still, the classifiers show varying performance on relevant benchmark datasets and, thus, potentially produce misclassifications. While our experiments show that the language with the strongest domain mismatches and weakest classifiers, Italian, has the lowest prediction change rates, we can not guarantee that the models do not misclassify some of the samples. Nevertheless, for our investigation, the actual label is not important as we are only interested in whether the labels for the original and simplified versions change. Even if the initial classification is wrong, consistent models should still produce the same label for all simplifications.

We observe alarmingly high prediction change rates across all languages and even OpenAI's GPT-3.5 model.
This suggests that the pre-training data of these models lack samples in simplified language and, thus, that there is still only little information available in plain language. Increasing internet accessibility and, hence, increasing the amount of data in simplified language is the most promising approach to improving plain language understanding in language models.
If such incoherence is not addressed, our findings empirically show that simplification can easily be exploited as a zero-iteration model-agnostic adversarial attack. Attackers would only need to simplify any input text to achieve success rates of up to 50\% with little effort.

\section{Conclusion}
In this paper, we have investigated how coherent models perform on simplified texts.
We have shown that different classifiers across multiple languages struggle with plain language samples and exhibit incoherent behavior. Such incoherency seems to affect also s.o.t.a LLMs like OpenAI's GPT3.5, which are not robust to simplifications from our benchmark datasets. We exploited human-created or human-aligned text simplification corpora to ensure natural and meaning-preserving samples. In this setting, we have observed prediction change rates of more than 40\%, indicating a severe lack of plain language understanding in pre-trained language models.

In future studies, we aim to expand our experiments to include more languages and settings involving automatically generated simplifications. Additionally, we believe that improving model coherence on simplified inputs can be achieved through human-preference tuning techniques—such as RLHF and DPO—and we encourage researchers to explore this direction further.

\section*{Ethical considerations}
We investigated whether model predictions change between original and simplified versions of the same text. Our findings can be valuable for identifying models' shortcomings and improving their robustness. However, our results can potentially be misused as adversarial attacks, and as such, they can threaten applications based on large language models. However, in our approach, we only re-use existing corpora and do not craft a new adversarial threat.

Especially for people with reading difficulties, the availability of information in plain language is crucial. Our experiments demonstrate that pre-trained language models are sensitive to plain language and that simplified samples are still underrepresented in pre-training corpora. This can cause severe problems when these people use LLM applications such as ChatGPT. Hence, we ask content creators and data scientists to remember the need for plain language and provide resources accordingly.

\section*{Data/Code availability statement}
Our work utilizes existing text simplification corpora. These corpora are publicly available except for the English Newsela corpus \citeplanguageresource{xu-newsela}. In addition, we publish our experiment code and links to these corpora at \href{https://github.com/MiriUll/LLM-consistency-simplification}{https://github.com/MiriUll/LLM-consistency-simplification}.

\nocite{*}
\section{Bibliographical References}\label{sec:reference}

\bibliographystyle{lrec-coling2024-natbib}
\bibliography{lrec-coling2024-example}

\section{Language Resource References}
\label{lr:ref}
\bibliographystylelanguageresource{lrec-coling2024-natbib}
\bibliographylanguageresource{languageresource}

\end{document}